\documentclass[sigconf]{acmart}
\AtBeginDocument{%
  }

\usepackage{graphicx}
\usepackage{subcaption}

\usepackage{booktabs}
\usepackage{siunitx}
\usepackage{array}

\usepackage{multirow, tabularx}
\usepackage{tcolorbox}
\tcbuselibrary{skins,breakable}

\usepackage{enumitem}

\begin{document}

%%
%% The "title" command has an optional parameter,
%% allowing the author to define a "short title" to be used in page headers.
\title{Efficient Personalization of Generative User Interfaces}

\author{Yi-Hao Peng}
\affiliation{%
  \institution{Massachusetts Institute of Technology}
  \city{}
  \country{}}
\email{yihao@csail.mit.edu}

\author{Jeffrey P. Bigham}
\affiliation{%
  \institution{Carnegie Mellon University}
  \city{}
  \country{}
}
\email{jbigham@cs.cmu.edu}

\author{Jason Wu}
\affiliation{%
 \institution{Purdue University}
 \city{}
 \country{}}
\email{jasonwu@purdue.edu}

\newcommand{\ipstart}[1]{\vspace{1mm}\noindent{\textbf{\textit{#1}}}}

\newcommand{\dataset}{DesignPref}
\newcommand{\datasetR}{DesignPref-R}

%%
%% By default, the full list of authors will be used in the page
%% headers. Often, this list is too long, and will overlap
%% other information printed in the page headers. This command allows
%% the author to define a more concise list
%% of authors' names for this purpose.

% Disable ACM reference format
% \settopmatter{printacmref=false}
\settopmatter{printacmref=false, printfolios=true}

% \settopmatter{authorsperrow=3}

% Remove copyright information
\setcopyright{none}

% Completely remove headers (conference name etc.)
\fancyhead{}

% Remove footnote for ACM rights text
\renewcommand\footnotetextcopyrightpermission[1]{}

\renewcommand{\shortauthors}{Peng et al.}

\keywords{Generative  UI, Personalization, Preference Learning, Alignment, Efficient Learning, Active Learning}

%%% Reinforcement Learning from Human Feedback

% personalization beyond prompting

%% A "teaser" image appears between the author and affiliation
%% information and the body of the document, and typically spans the
%% page.

% \begin{teaserfigure}
%   \includegraphics[width=\textwidth]{sampleteaser}
%   \caption{Seattle Mariners at Spring Training, 2010.}
%   \Description{Enjoying the baseball game from the third-base
%   seats. Ichiro Suzuki preparing to bat.}
%   \label{fig:teaser}
% \end{teaserfigure}

% \received{20 February 2007}
% \received[revised]{12 March 2009}
% \received[accepted]{5 June 2009}

%%
%% This command processes the author and affiliation and title
%% information and builds the first part of the formatted document.

\begin{abstract}

Generative user interfaces (GenUIs) create new opportunities to adapt interfaces to individual users on demand. Yet personalization is difficult because it is not possible to provide settings for screens that have not yet been generated, making it necessary to learn preferences from users' feedback on generated interfaces. Such feedback is sparse, subjective, and often difficult to articulate. We study this problem through a new dataset in which 20 participants each judge the same 600 pairs of GenUIs. Their judgments differ substantially (Krippendorff's $\alpha=0.25$), and written rationales show that even when participants consider similar UI properties, such as color tone or hierarchy, they often make different choices and prioritize them differently. We develop a sample-efficient personalization method that uses a few pairwise judgments to weight prior users' preferences rather than relying on a fixed rubric of UI attributes. Our method outperforms a pretrained UI evaluator and a larger multimodal model offline. In an online study with 12 new users, participants preferred interfaces personalized by our method over all baselines, including shared guidelines and users' own written preferences.

\end{abstract}

\begin{teaserfigure}
  \centering
  \includegraphics[width=0.97\textwidth]{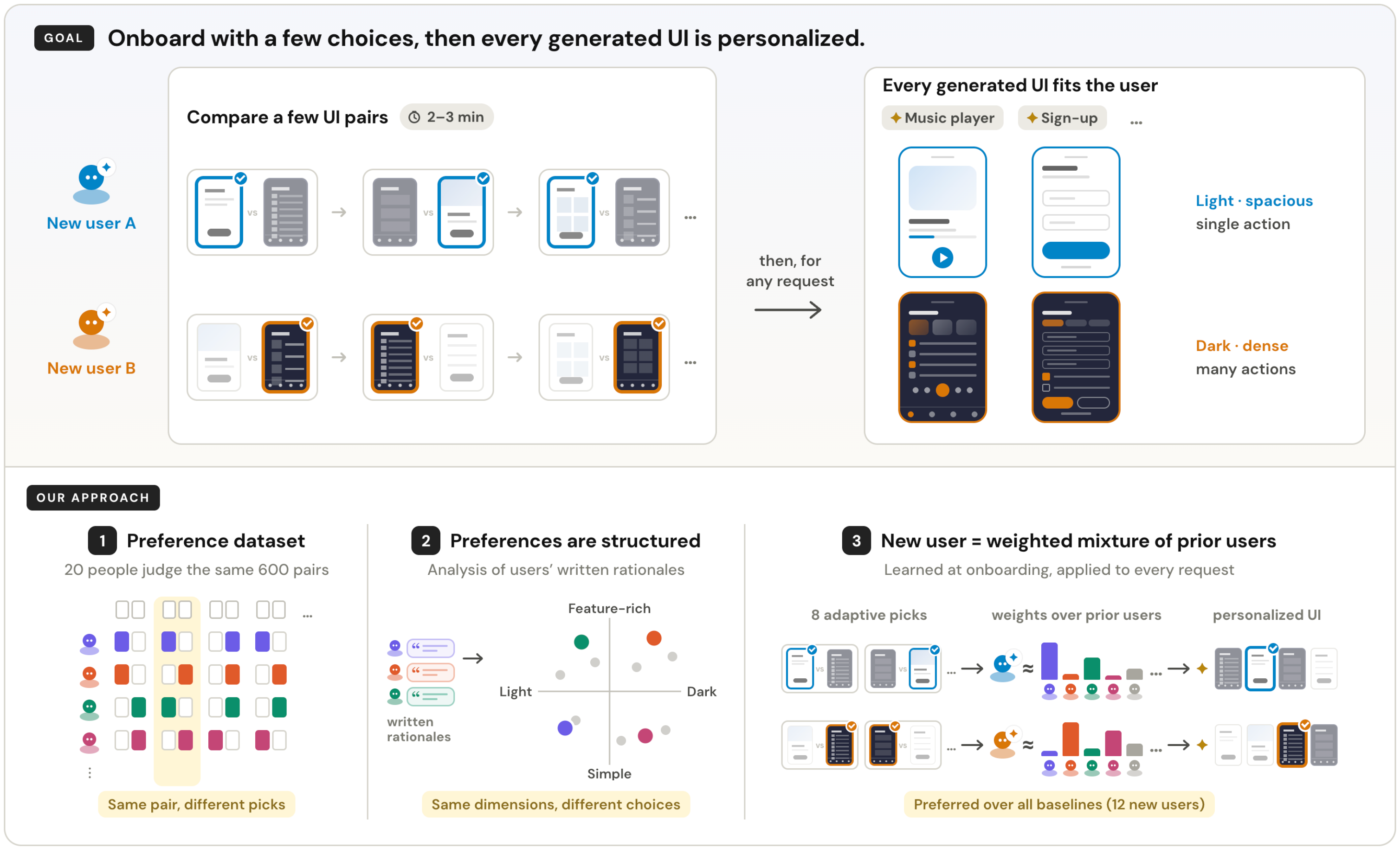}
  \caption{Overview of our methods towards personalizing generative UIs from a few choices. Top (goal): A new user answers a few choices during a brief 2-3 minute onboarding. For subsequent requests, the system selects generated UIs that better match the user's preferences, so the same request can yield different interfaces for different users. Bottom (our approach): 1) We collect 12,000 judgments from 20 people who evaluate the same 600 UI pairs and observe frequent disagreement. 2) Choice patterns and written rationales reveal structure in this disagreement: people attend to similar design dimensions, such as theme and layout complexity, but differ in their preferred values and relative importance. 3) We model a new user as a weighted mixture of prior users, infer the weights from eight adaptively selected comparisons, and use the learned mixture to rank candidate UIs for each new request.} 
  \label{fig:teaser_pipeline}
  \Description[]{}
  \vspace{16pt}
\end{teaserfigure}

\maketitle

\section{Introduction}
\label{sec:intro}

Personalizing user interfaces (UIs) for individual users is a long-standing goal in human-centered computing. Prior work has adapted interfaces to measurable user characteristics such as abilities and device constraints~\cite{gajos2010supple,wobbrock2011ability}, as well as factors such as task goals and interaction context~\cite{paterno1997concurtasktrees,horvitz1999principles}. Yet many interface preferences remain difficult to model because they are subjective and hard to articulate.

Generative user interfaces (GenUIs) make this challenge especially salient. Rather than selecting from a fixed set of pre-authored designs, GenUIs can be synthesized on demand to fulfill a user's request~\cite{moran2024generative,chen2025generative}. This flexibility creates new opportunities for personalization, since UI properties such as layout and color can be adapted at generation time. However, because such interfaces may be ephemeral and created for a single interaction, persistent settings or a small set of handcrafted variants may capture only part of a user's preferences. The preferences that matter for generation are also difficult to specify in advance, whether through a fixed set of generator controls or instructions in a system prompt~\cite{gajos2005arnauld,alves2026interactiondata}. We envision GenUI personalization that learns from a user's prior choices and applies those preferences to future requests. Users should not need to specify the same interface preferences each time a new interface is generated (Figure~\ref{fig:teaser_pipeline}).

Such personalization must also work when little interaction history is available. As in recommender systems, new users present a cold-start problem. For practical use, onboarding should require only a few explicit choices and take little time. YouTube Music, for example, asks new users to select five preferred artists to initialize personalization~\cite{hsu2024minimizing}. Because each GenUI request may produce new candidate interfaces, the model must generalize beyond the examples used to learn the user's preferences. Moreover, if interface preferences differ systematically across people, a single aggregate model may miss differences that matter to an individual user. A practical method could therefore draw on preference data from prior users while adapting to each new user with limited feedback.

In this paper, we study whether we can efficiently model individual users' preferences for generative UIs and use them to personalize generated interfaces. We collect a dataset in which 20 people each judge the same 600 pairs of generated UIs, which allows us to compare preferences directly across users. Their judgments differ substantially (Krippendorff's $\alpha=0.25$). Written rationales reveal recurring considerations, such as color theme, imagery, and readability, but participants describe different preferred values and place different emphasis on those considerations. For example, some favor light themes or simple layouts, while others favor dark themes or more complete interfaces; some dimensions also matter more to some users than to others.

Motivated by these findings, we develop a sample-efficient personalization method that represents a new user's preferences as a weighted mixture of prior users' preferences rather than a fixed rubric of UI properties. A few adaptively selected pairwise comparisons update the weights over prior users. For a new UI pair, the model retrieves visually similar comparisons from the preference dataset and combines their labels according to these learned user weights. In offline evaluation, our method achieves the highest preference prediction accuracy across the evaluated feedback budgets, outperforming baselines that include collaborative filtering, direct model adaptation, and  larger multimodal models. We then integrate the learned preference model into an end-to-end personalized generation pipeline, where it reranks candidates from a base generator. In an end-to-end study with 12 new users, participants preferred interfaces personalized by our method over all baseline approaches, including generation conditioned on users' self-described preferences and generic UI design guidelines.

In summary, our work contributes: 1) A new dataset for studying personalization in generative UIs, consisting of 12,000 pairwise preference labels from 20 people over 600 comparison pairs; 2) An analysis of preferences in generative UIs, showing substantial disagreement across people and differences in how people choose and prioritize UI properties; and 3) A sample-efficient personalization method that outperforms existing baselines in both offline preference prediction and online interface generation.
%
% In summary, this paper makes three contributions:
% \begin{itemize}
%     \item We introduce a new dataset for studying personalization in generative UIs, consisting of 12,000 pairwise preference labels from 20 users and 600 generated interfaces.
%     \item We provide a data-driven analysis of preference divergence in interface design, showing substantial disagreement across users and qualitative differences in how shared design concepts are interpreted.
%     \item We apply our findings to develop a sample-efficient personalization method and end-to-end personalized generation pipeline that outperform strong baselines in both offline preference prediction and online interface generation.
% \end{itemize}
%

% We plan to release our data and models upon acceptance.

% \vspace{-5pt}

% \section{Related Work} 
% \label{sec:related_work} 
% Our work sits at the intersection of UI evaluation, UI adaptation, and preference learning. These areas cover how to judge interface quality, how to adapt interfaces to broad classes of user needs, and how to learn from human preference data. Our setting differs in that a generative UI system often needs to choose among dynamically created designs for a specific person before much interaction history or many labels are available.

% \section{Related Work}
% \label{sec:related_work}
% Our work on personalized preference alignment for generative UIs builds on three research areas, (i) UI evaluation, (ii) UI adaptation, and (iii) preference learning and personalization.

\section{Related Work}
\label{sec:related_work}
Our work builds on prior research on how interfaces are assessed, adapted, and improved from human feedback. We review three areas most relevant to our setting: 1) UI evaluation, 2) Adaptive and generative UIs, and 3) Efficient user preference learning.

\subsection{UI Evaluation} 
Early computational approaches to UI evaluation focused on properties that most people tend to agree on. Hand-crafted measures for aesthetics, clutter, and complexity were later packaged into reusable perceptual analysis tools~\cite{miniukovich2015aesthetics,miniukovich2018complexity,oulasvirta2018aim}. Affective and physiological signals offered a lower-effort way to distinguish stronger from weaker interfaces~\cite{haddad2024goodguis}. These methods are useful for measuring common perceptual qualities but do not model how taste differs across individuals. More recent work replaces fixed metrics with learned critics. Prompting large multimodal or vision-language models (LMM/VLM) with design heuristics can produce useful critiques of UI mockups, and pairing users' annotations with region-level quality scores strengthens that direction~\cite{duan2024generating,duan2024uicrit}. UI-specific evaluators trained for design quality and prompt relevance push evaluation further, though benchmarks show that general-purpose multimodal models still match human UI judgments only moderately well~\cite{wu2024uiclip,luera2025mllm}. Other work broadens the scope of evaluation itself. Some tie pairwise assessment to real A/B test outcomes, while others show that richer feedback modalities like comments, sketches, and direct manipulation can supervise UI generation more effectively than rankings alone~\cite{jeon2026wiserui,wu2025designerfeedback}. Our work complements general UI assessment with an approach that adapts preference predictions to an individual user.
% All of these efforts aim at a single shared standard of quality. Our work instead focuses on personalized judgment, studying how an evaluator can adapt to a particular user's preferences while preserving shared notions of UI quality.

% \vspace{-5pt}

\subsection{Adaptive and Generative User Interfaces}
Prior work adapts interfaces to users through predefined objectives, rules, examples, or explicit user input. SUPPLE searches over interface configurations using objective functions that encode device constraints, user abilities, and preferences~\cite{gajos2004supple,gajos2007supplepp,gajos2010supple}. Mixed-initiative and adaptable systems instead let users inspect, reject, or directly specify interface changes~\cite{bunt2007mixedinitiative,proenca2021uiflex,alves2024citizenled}. FrameKit uses author-provided examples to define how an interface can vary across contexts~\cite{wu2024framekit}. These systems support adaptation within author-defined configuration spaces or examples.
Generative UIs broaden this space through new interface structures and presentations at runtime~\cite{chen2025generative,cao2025generative}.
Recent systems let users reshape generated interfaces through natural language, direct manipulation, or structured specifications~\cite{cao2025generative,min2025overviewdetail,chen2025specifyui,peng2025morae}. AlignUI further uses prior preference data to generate controls that match a stated user preference~\cite{liu2026alignui}. We focus on preferences that users may not know how to specify before seeing alternatives. Our method learns these preferences from a few choices between complete generated interfaces and uses them to personalize new generation.

\subsection{Efficient User Preference Learning}

Preference elicitation asks how a system can infer what a user values from a small number of interactions. In HCI, systems like ARNAULD learn weights over interface costs from users' choices between alternatives~\cite{gajos2005arnauld}. Choice-based recommender systems similarly combine a few representative choices with preference data from prior users to infer a new user's preferences~\cite{loepp2014choice}. These approaches motivate pairwise comparisons as a lightweight way to personalize when little user-specific data is available.

How the system represents a user determines what it can learn from those few choices. AMPLe estimates a user's weights over predefined preference dimensions and actively selects informative comparisons~\cite{oh2024ample}. PAL instead learns shared prototypes and adapts a user's mixture over them from a few examples~\cite{chen2025pal}. Our method represents a new user as a mixture over prior users' choices. Adaptive comparisons update the user weights, while visual retrieval identifies which prior UI judgments are relevant to a new pair. This avoids requiring either a fixed rubric of UI attributes or a single matching user.

The learned preferences must also transfer beyond the interfaces shown during onboarding. Choice-based collaborative filtering learns from prior users' item preferences~\cite{loepp2014choice}, while visual methods such as VBPR use image features to generalize across items~\cite{he2016vbpr}. Generated UI candidates in our setting rarely recur, so we retrieve visually similar pairwise comparisons and combine their labels using the learned user weights. UserAlign also elicits preferences through a few comparisons, but uses them to select among candidates for a particular request~\cite{padurean2025useralign}. We instead study whether a brief onboarding can support preference prediction across new unseen UI prompts, without requiring users to repeat preference elicitation for each request.

\section{Preference Dataset for Generative UIs}
\label{sec:dataset}

To personalize generative UIs for individual users, we first need to understand how people evaluate generated interfaces and how much their preferences differ. We collect pairwise preference judgments from 20 participants over the same 600 pairs of generated mobile UIs, resulting in 12,000 preference labels.
We use pairwise comparisons because they require less cognitive effort than absolute ratings and can elicit preferences that users find difficult to articulate directly~\cite{gajos2005arnauld,kalloori2018pairwise}.
We focus on mobile UIs because existing datasets provide broader coverage of this domain, but the same approach could apply to other interface and visual design domains, including webpages, slides~\cite{ge2025autopresent,peng2023slide,yun2025designlab,peng2022diffscriber}, and posters~\cite{hsu2023posterlayout,hsu2025postero}.

\subsection{Dataset Construction}
\label{sec:designpref}

Existing UI datasets provide screenshots, text descriptions, and other metadata but do not include annotations tied to individual preferences~\cite{wang2021screen2words,bunian2021vins,wu2024uiclip,duan2024uicrit}. We use human-written screen descriptions from these datasets as prompts and generate UIs with GPT-5 and Gemini-2.5-Pro, two high-ranked models on WebDev Arena~\cite{webdev_arena_2025} at the time of data collection. The models generate HTML, CSS, and JavaScript that we render as mobile UI screens.

\ipstart{UI Generation.}
We sample 100 human-written screen descriptions from an existing UI summary dataset~\cite{wang2021screen2words} and expand each into a more complete prompt using established methods~\cite{peng2024dreamstruct,wu2024uicoder}. To cover a range of UI types, we balance the prompts across categories from an existing taxonomy~\cite{leiva2020enrico,wu2023webui}. Across the 100 prompts, GPT-5 and Gemini-2.5-Pro produce 4 variants per prompt at high temperature to increase diversity. We follow Design Arena~\cite{design_arena_2025}'s system prompts to produce high-quality, renderable UI code and include explicit quality checks in the instructions. In total, the dataset contains 400 generated UIs. Each prompt yields six pairwise combinations among its four variants, resulting in 600 unique comparison pairs.

\ipstart{Annotation Protocol.}
We recruited 20 participants through a mailing list and online platforms. Participants had varied levels of UI design experience, ranging from current university students who had completed at least one introductory interface design course to people whose jobs involved design-related tasks.
To reduce fatigue, participants could pause and resume the study across multiple sessions. The study interface required participants to stop the session timer when leaving and restart it when returning, and we summed active time across sessions. Participants spent approximately 2.5--4 hours actively completing the comparisons. Participants spent somewhat longer per comparison than in recent UI comparison studies~\cite{wu2025designerfeedback,perraud2026interface}.
Our four-point scale captured both the direction and strength of preference.
In each trial, a participant viewed a pair of UI screens $(A,B)$ and selected one of four options: ``A much better than B'' ($A\gg B$), ``A better than B'' ($A>B$), ``B better than A'' ($B>A$), or ``B much better than A'' ($B\gg A$). We used an even-numbered scale with no neutral option to require a directional choice while retaining preference strength. Similar fine-grained preference scales have been used in model alignment and preference evaluation~\cite{touvron2023llama2,song2024veriscore,wang2024lrhp}. We retained a participant identifier for each judgment so that we could model and compare preferences at the individual level, following prior work on annotator-specific modeling in crowdsourcing~\cite{chen2013pairwise,caron2012gbt,dawid1979ml,raykar2010crowds}. Participants were compensated \$15.55/hour, and our institution's IRB approved the study.

% \begin{figure*}[t]
%     \centering
%     \includegraphics[width=0.45\textwidth]{figures/figure3a.pdf}\hfill
%     \includegraphics[width=0.45\textwidth]{figures/figure3b.pdf}
%     \caption{User-level structure in preference rationales. Numbers identify participants. Panel (a) summarizes imagery and evaluation emphasis on 53 contested comparisons; panel (b) summarizes presentation and theme on the other 53. Positions reflect the balance of coded explanations.}
%     \Description{Two scatterplots with 13 numbered participants in each. Panel (a) places imagery against evaluation emphasis. Participants 6 and 7 both have positive imagery scores, but participant 6 emphasizes readability and clarity and participant 7 emphasizes appearance and polish. Panel (b) places simplicity versus completeness against light versus dark theme. Participants 10 and 17 differ on presentation despite both having positive imagery and appearance scores in panel (a). Participants 3, 10, and 17 have no coded theme direction, which appears at zero.}
%     \label{fig:rationale_structure}
% \end{figure*}

\begin{figure}[!htp]
    \centering
    \includegraphics[width=0.45\textwidth]{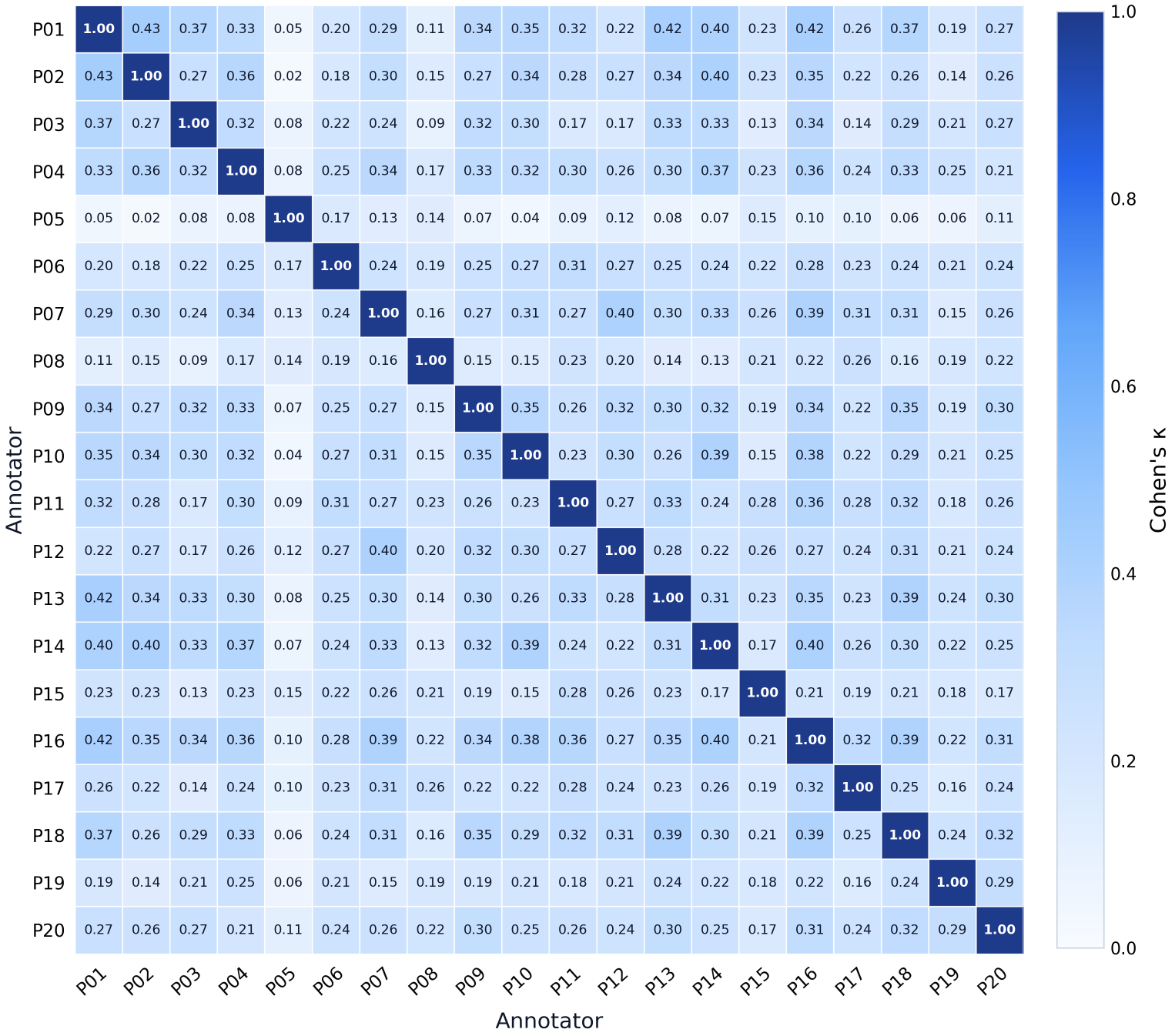}
    \vspace{-5pt}
    \caption{Cohen’s kappa scores for pairwise binary preference agreement between participants (average $\kappa=0.25$).}
    \vspace{-12pt}
    \label{fig:kappa}
\end{figure}

\subsection{Preference Label Analysis}
\label{sec:dataset_analysis}

We analyze preference agreement across participants at two levels of granularity: a binary label that records which variant wins and a four-way label that encodes preference strength. Generally, people do not agree much at either level. For binary preferences, mean pairwise agreement is $0.624$, with Cohen's $\kappa$ and Krippendorff's $\alpha$ both at $0.248$. For nominal four-way labels, agreement drops to 0.386, $\kappa$ to 0.114, and $\alpha$ to 0.104. Multi-rater Fleiss' $\kappa$ is 0.248 for binary choices, and ordinal Krippendorff's $\alpha$ is 0.265 for the four-way scale, so agreement remains limited when the analysis accounts for preference order. The agreement rates align with prior UI preference and critique studies~\cite{wu2024uiclip,duan2024uicrit}. Direction and four-way $\kappa$ correlate at $r=0.60$, so pairs of participants who agree on which variant wins also tend to agree on how strongly the variant wins, though strength judgments introduce additional variation. Part of the gap between direction and strength reliability comes from how participants use the scale. Across all judgments, $75.54\%$ fall in the middle categories (A$>$B or B$>$A) and $24.46\%$ in the extreme categories (A$\gg$B or B$\gg$A). But the split varies widely across participants: the middle-option share ranges from $6.00\%$ to $99.50\%$, and the extreme-option share from $0.17\%$ to $93.67\%$. Participants set different bars for what counts as a strong preference. Weak choices match the majority choice on 66.9\% of judgments, compared with 77.7\% for strong choices. 
% This association does not establish that any particular weak choice was confident. Low agreement alone cannot distinguish personal taste from uncertainty, so Section~\ref{sec:rationales} examines whether participants' explanations contain recurring differences.

\ipstart{Effect of Screen Type.}
Some UI screen types leave more room for variation than others. For example, a social media feed allows many plausible layouts, while a login screen is more constrained. To examine whether screen type affects agreement, we group comparison pairs according to a prior taxonomy~\cite{leiva2020enrico} and measure direction consensus through per-item mean pairwise agreement and the entropy of binary votes, averaged within each type. \emph{Form} screens show the strongest consensus, with mean pairwise agreement of $72.11\%$ and the highest share of extreme choices at $32.71\%$. \emph{List}, \emph{Dialer}, and \emph{Search} screens follow with agreement around $66\%$--$68\%$. Users disagree most on \emph{Terms} pages, \emph{Camera} interfaces, \emph{Media Player} screens, and \emph{Login} flows, all near $54\%$--$57\%$ agreement. Across the 20 screen types, mean agreement and the share of extreme judgments correlate strongly ($r=0.774$). Screens with structured layouts and clear primary actions (e.g., forms, search pages) tend to produce more unified judgments, while visually complex or multifunctional screens lead to more diverse preferences.

\subsection{Preference Rationales and Structures}
\label{sec:rationales}

\begin{figure}[!htp]
    \centering
    \includegraphics[width=0.475\textwidth]{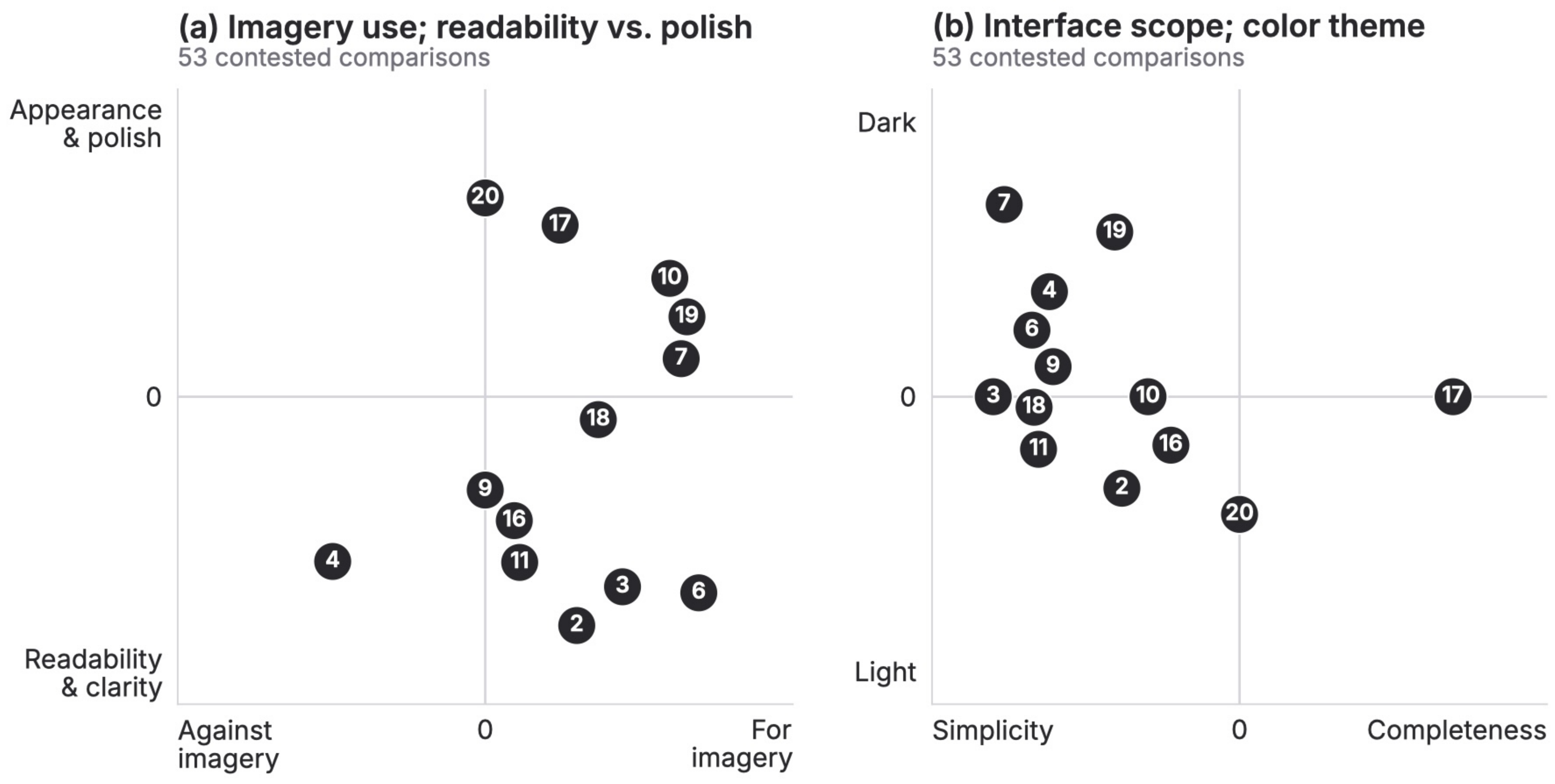}
    \vspace{-10pt}
    \caption{User-level structure emerged in preference rationales across contested UI comparisons. Numbers denote participant IDs. Panel (a) plots imagery use against readability versus visual polish using rationales from 53 contested comparisons; panel (b) plots interface scope (simplicity versus completeness) against color theme using rationales from the other 53. Positions reflect each participant's balance of coded explanations.}
    \vspace{-9pt}
    \label{fig:rationale_structure}
\end{figure}

To understand the reasons behind people's choices, we invited the same 20 participants to a follow-up rationale study; 13 participated. We selected the 106 pairs with highly divergent preferences, defined as 10--10 or 11--9 vote splits among the 20 participants. For each pair, participants saw their earlier choice, wrote a brief rationale, and could revise their choice; 92.7\% of choices remained unchanged. The study yielded 1{,}378 rationales and took about 3.5--5 hours per participant. We examine both recurring patterns across the rationales and the combinations of considerations that distinguish users (Figure~\ref{fig:rationale_structure}).

\ipstart{Analysis Method.}
We analyze the rationales in three stages. First, we split each rationale into sentences, embed the sentences with a transformer-based encoder~\cite{reimers2019sentence}, and cluster similar comments. Following prior work on LLM-assisted thematic analysis~\cite{qiao2025thematic,wang2025lata,beeferman2023feedbackmap}, GPT-5 assigns each cluster a neutral label and summary, then merges related clusters. Second, for each contested pair, we group rationales by the chosen screen and compare the reasons on the two sides. This step identifies cases where users discuss the same UI dimension but prefer different values within it. Third, we build a fixed dictionary from recurring UI terms and their contexts to compare patterns across users. We initialize the dictionary by Ward-clustering recurring terms using term and context embeddings, remove content-specific terms, and merge related groups. Four main considerations recur in the pair-level analysis and support consistent coding across users: imagery use; readability versus visual polish; interface scope, from simplicity to completeness; and color theme, from light to dark. For imagery use and color theme, references to chosen and rejected screens, negative language, and absence terms determine direction. For readability versus visual polish and interface scope, the score records which concept a rationale emphasizes. Each rationale contributes at most one count to either side of a dimension. For each user, we compute a symmetrically smoothed contrast, $(n^+-n^-)/(n^++n^-+2)$~\cite{bishop2006pattern}. The equivalent add-one smoothing assigns one pseudocount to each pole, which shrinks estimates supported by few coded rationales toward zero. Figure~\ref{fig:rationale_structure} shows how these dimensions vary across users using two non-overlapping sets of 53 contested pairs. One highlights imagery use and readability versus visual polish, while the other highlights interface scope and color theme.

% Figure~\ref{fig:rationale_structure} uses separate 53-pair display sets (half of the whole set) for its two panels.

\begin{figure*}[!htp]
    \centering
    \begin{subfigure}[htbp]{0.495\textwidth}
        \centering
        \includegraphics[width=\linewidth]{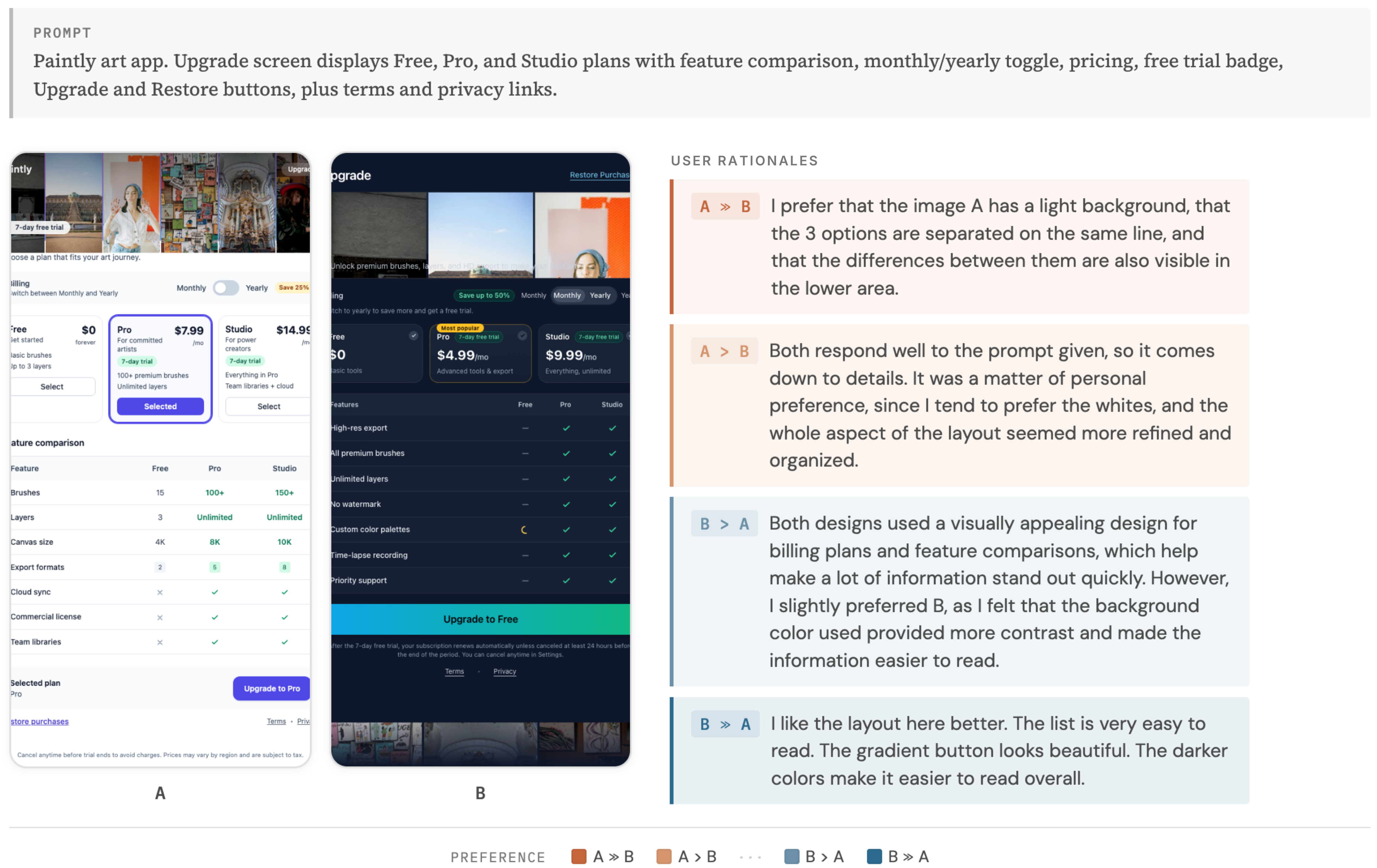}
        \caption{In the Upgrade screen, users diverge in preference for light versus dark themes and contrast.}
        \label{fig:div_exmp_1}
    \end{subfigure}
    \hfill
    \begin{subfigure}[htbp]{0.495\textwidth}
        \centering
        \includegraphics[width=\linewidth]{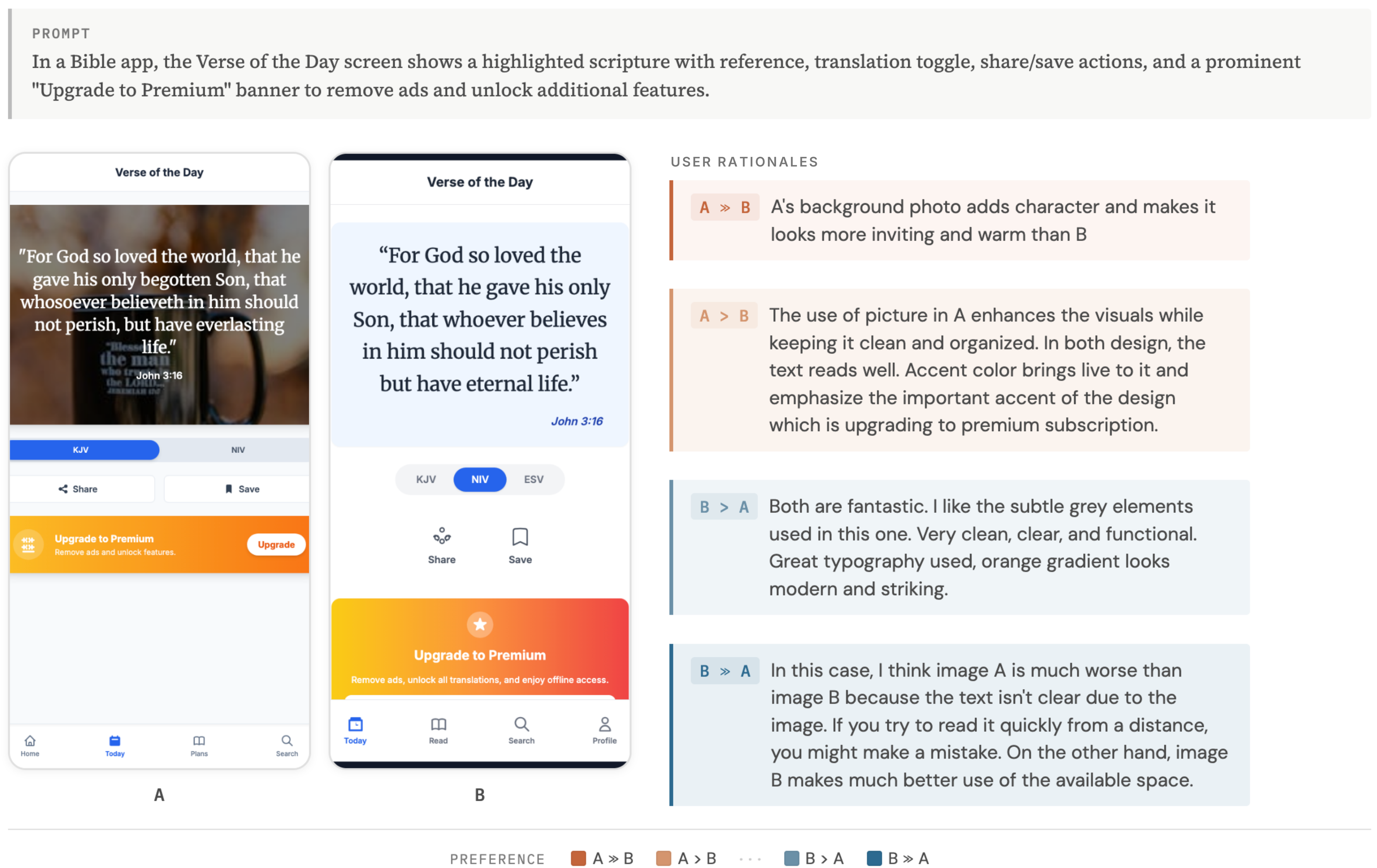}
        \caption{In the Bible app screen, users diverge between decorative imagery and a more utility-focused layout.}
        \label{fig:div_exmp_2}
    \end{subfigure}
    \vspace{-7.5pt}
    \caption{Rationales for divergent preferences across users. Each comparison shows an even preference split between A and B.}
    % \vspace{-7.5pt}
    \label{fig:div_exmp}
\end{figure*}

\ipstart{Findings.}
We found that users repeatedly mentioned the four extracted aspects while the comments reveal two forms of divergence. Users can prefer different values within a dimension, such as light versus dark themes, more versus less imagery, or simpler versus more complete interfaces. Users also differ in which dimensions they emphasize or prioritize, such that one user may emphasize readability while another emphasizes visual polish, and some dimensions may not appear in a user's explanation at all.

Figure~\ref{fig:div_exmp} shows the first form of divergence in two actual pairs. On the Upgrade screen, users split between light and dark visual treatments; on the Bible app screen, users split between decorative imagery and a more utility-focused layout. Figure~\ref{fig:rationale_structure} shows the second form at the user level. P06 and P07 both lean toward imagery use, but P06 more often emphasizes readability and clarity while P07 emphasizes appearance and polish. P10 and P17 similarly lean toward imagery use and visual polish, yet P10 favors simpler interface scope while P17 favors completeness. A user's preference on one dimension therefore gives only partial information about the user's broader preference pattern.

We test whether these user-level differences recur across prompts. Across 500 random splits of all 63 prompts represented in the rationale study, we keep every pair from a prompt in the same half and recompute each user's scores. Readability versus visual polish and interface scope have median cross-half Pearson correlations of $r=0.92$ and $0.84$, with a median of 13 eligible participants. Imagery use and color theme show weaker recurrence ($r=0.68$ and $0.53$) and less coverage (11 and 7 participants at the median). For each dimension, eligibility requires coded evidence from at least two prompts in each half. The dictionary remains fixed across splits after development on the full corpus. These correlations describe the level of high recurrence in coded explanations.

% \ipstart{Implications for Modeling.}
% \label{sec:modeling_motivation}
% The rationale analysis suggests a couple of requirements for personalization. Recurring dimensions across users provide shared preference signal. Different preferred values and priorities make a single aggregate preference function insufficient. Partial overlap across dimensions also makes a single prior user or fixed user type too restrictive. For instance, users who prefer UIs with imagery may differ in their preferences for readability or color themes (Figure~\ref{fig:rationale_structure}). The relevant dimensions can also vary across interface pairs. Some comparisons foreground color theme, while others expose differences in imagery or readability. A preference model should therefore account for both which prior users provide useful signal and which past UI comparisons are relevant to the current pair.
% Motivated by our findings, we present a method that represents a new user as a soft distribution over prior users and updates the distribution from a few onboarding comparisons. For each new UI pair, we retrieve visually similar past comparisons and combine their labels according to the learned user weights. This formulation reuses prior users' preferences without reducing each user to a fixed rubric or hard cluster.

\ipstart{Implications for Modeling.}
\label{sec:modeling_motivation}
Our analyses suggest that personalization requires preserving differences between users rather than collapsing their judgments into a single notion of interface quality. First, the low agreement in our dataset shows that aggregating a larger volume of preference labels may obscure meaningful variation that is needed for accurate preference signals. Our rationale analysis revealed some structured factors in this disagreement. First, different participants tended to use different dimensions (e.g., color, layout, font) from one another, suggesting that it is difficult to establish a global rubric. In addition, even participants who focused on similar axes, they often preferred tradeoff directions, e.g., preferring images as a visualization aid vs rejecting images as distractions (Figure~\ref{fig:div_exmp}). These context-dependent judgments also complicate how user opinion can both be elicited and used for evaluation.
Finally, the dimensions that participants cited were dependent on the comparison being judged, e.g., more likely to bring up font readability for a reading app. This makes preferences difficult to specify in advance (e.g., as a user-authored persona), since which preferences matter often becomes apparent only in the context of a particular interface.

We hypothesize that an effective personalization model would be able to directly draw on previous examples, giving the adequate context for evaluation. With sufficient interaction history, these signals could be learned from each user independently. However, in most end-user settings, such as our described on-boarding scenario, each user would only provide a few judgments before use. We therefore hypothesize that prior feedback from similar users can compensate for sparse individual history and use this to drive our modeling experiments.

\section{Preference Modeling}
\label{sec:modeling}

Based on the insights of our dataset analysis, we introduce a sample-efficient personalization method for adapting a preference model to a new user from only a small number of observations. Our goal is to ask the user a limited set of comparison queries and use their responses to infer weights over users in the training set.
To adapt from sparse feedback, we represent the new user as a mixture of similar users' choices in the training set, for whom sufficient labeled data is available to support effective data-driven modeling, and use this mixture to personalize predictions. Our offline evaluation compares this approach with direct parameter adaptation, collaborative filtering, and multimodal prompting.

% We introduce a sample-efficient personalization method for adapting a preference model to a new user from only a small number of observations. Our goal is to ask the user a limited set of comparison queries and use their responses to infer a prior over users in the training set.
% %
% Because data from a new user is scarce, directly fine-tuning a model is not practical. Instead, we represent the new user as a mixture of similar users in the training set, for whom sufficient labeled data is available to support effective data-driven modeling, and use this mixture to personalize predictions. Our offline evaluation shows that this approach outperforms a substantially larger multimodal model (GPT-5.2), both in absolute performance and in its ability to benefit from additional user queries.

\subsection{Modeling Approach}
We describe a retrieval-based approach to preference modeling in which retrieval from a labeled data bank is personalized using onboarding pairs from a new user.
The method consists of three stages: i) data bank construction, ii) query selection, and iii) model personalization.

\ipstart{Data Bank Construction.}
We first convert the collected data into a consistent representation for modeling.
Let \(x = (t, I^A, I^B)\) denote a screen pair, where \(t\) is the text prompt and \(I^A\) and \(I^B\) are the two candidate screens.
Each labeled example is then represented as
\begin{equation}
\mathcal{D}
\;=\;
\{\, (x, d, c) \;:\; c \in \{-2,-1,1,2\} \,\},
\label{eq:preference-data}
\end{equation}
where \(d\) is an anonymized user identifier and \(c\) is the recorded preference label.
We also construct stratified train, validation, and test splits using a \(60/20/20\) ratio.
All pairs containing a given screen ID are assigned to the same split to prevent data leakage.
This yields 7200 training samples and 2400 samples each for validation and testing.
For each user, this corresponds to 360 training samples and 120 validation and test samples.

\begin{figure}[!t]
    \centering
    \includegraphics[width=0.44\textwidth]{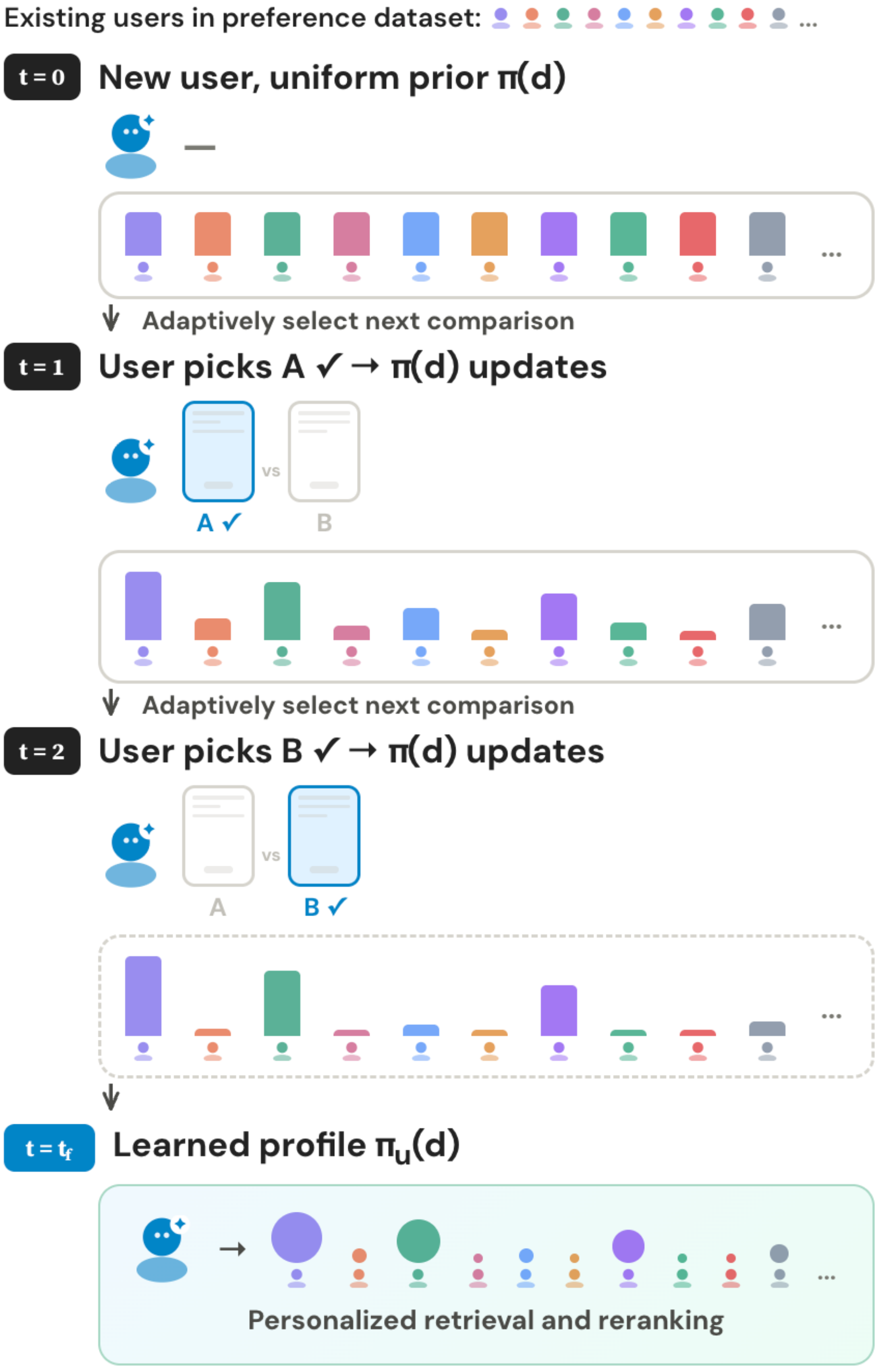}
    % \caption{Our query selection algorithm queries a new user with a small number of queries. These queries are used to learn a ``profile'' of the user, which is defined as a distribution over users in the training set. After each user response, the entropy of the distribution $\pi_t$ decreases, as probability mass concentrates on users whose preferences are more consistent with the new user's responses.}
    \caption{Our query selection algorithm asks a new user a small number of comparison queries. The responses define a user ``profile'' as a distribution $\pi_t$ over users in the training set. Each response updates this distribution, and the query policy selects comparisons that minimize the expected entropy of the updated distribution.}
    \vspace{-15pt}
    \label{fig:queryselection}
\end{figure}

\ipstart{Query Selection.}
Given the training split \(\mathcal{D}_{\mathrm{train}}\), we select up to \(k\) screen pairs to ask a new user during onboarding.
Each candidate query is a pair \(x\) drawn from the training bank.
We choose the next query using expected information gain (EIG), which favors queries that are expected to be most informative about which training users the new user most resembles~\cite{bickfordsmith2023prediction}.
Our approach is illustrated in Figure~\ref{fig:queryselection}.
We begin with a uniform prior over training users and update it after each user response.
Let
\[
\mathcal{O}_t = \{(x_1,c_1), \dots, (x_t,c_t)\}
\]
denote the onboarding responses observed after \(t\) queries, where \(c_i\) is the new user's response to pair \(x_i\).
We summarize the new user's current profile by a distribution \(\pi_t\) over training users, where \(\pi_t(d)\) is the current weight assigned to training user \(d\).
At \(t=0\), we initialize \(\pi_0\) to be uniform.
We score each candidate query \(x\) in the training split by its expected information gain.
To do so, we first compute the predictive distribution over possible responses under the current user distribution \(\pi_t(d)\):
\[
p(c \mid x,\pi_t)
=
\sum_d \pi_t(d)\, p(c \mid x,d),
\]
where \(\pi_t(d)\) is the current probability assigned to user \(d\), and \(p(c \mid x,d)\) is the probability that user \(d\) would give response \(c\) to pair \(x\).
If the new user answers query \(x\) with response \(c\), we update the user distribution \(\pi_t(d)\) by Bayes' rule:
\[
\pi_{t+1}(d)
=
\frac{
\pi_t(d)\, p(c \mid x,d)
}{
\sum_{d'} \pi_t(d')\, p(c \mid x,d')
}.
\]
Here, \(\pi_t(d)\) is the current probability assigned to training user \(d\), and \(p(c \mid x,d)\) is the probability that user \(d\) would give response \(c\) to pair \(x\).
This update increases the weight of users whose predicted response is more consistent with the observed answer.
We quantify uncertainty in the user profile by the entropy \(H(\pi_t)\) of the current user distribution, which measures how broadly probability is distributed across candidate users.
For a candidate query \(x\), we estimate the expected entropy after asking that query by averaging over possible user responses:
\[
\sum_{c \in \{-2,-1,1,2\}}
p(c \mid x,\pi_t)\, H(\pi_{t+1}).
\]
The expected information gain of \(x\) is the difference between the current entropy and the expected post-query entropy.
We select the candidate query with the largest EIG and present it to the user. This process is repeated $k$ times where $k$ is the number of onboarding pairs.

\ipstart{Model Personalization.}
After collecting onboarding responses from a new user \(u\), we use them to personalize a retrieval-based preference model.
Given a query pair \(q=(I_{\mathrm{left}}, I_{\mathrm{right}})\), the model retrieves similar labeled pairs from the training bank and aggregates their labels using the learned user distribution \(\pi_u\).
Each screenshot is embedded into a learned feature space by an encoder \(f_I\).
Because UIClip is trained specifically for GUI screenshots, we use its frozen visual encoder with a learnable residual layer on top.
For a query pair \(q\), we form a pair representation from the two screenshot embeddings and retrieve the top \(n\) nearest training pairs for soft-averaging by similarity. We chose \(n=100\) by manual tuning.

To compare ordered pairs, let \((z_L, z_R)\) denote the embeddings of the left and right images in the query pair, and let \((z_{j,L}, z_{j,R})\) denote the embeddings of bank pair \(j\).
We score similarity using both direct and swapped alignments,
\[
s(q,j) = z_L^\top z_{j,L} + z_R^\top z_{j,R},
\qquad
s'(q,j) = z_L^\top z_{j,R} + z_R^\top z_{j,L},
\]
and combine them with a soft maximum, which allows the model to handle pair symmetry.
Each of the $n$ retrieved pairs has multiple labels from the training users, and so we aggregate the labels based on the learned user distribution and then aggregate the pairs based on the pair similarity metric.
Let \(w_{qj}\) denote the retrieval weight for bank pair \(j\) under query \(q\).
For a new user \(u\), the onboarding stage produces user weights \(\pi_u\) over historical users.
We score a query pair \(q\) by combining the labels of retrieved bank pairs \(j\), weighted both by retrieval similarity and by user similarity:
\[
s(q \mid u) = \sum_j w_{qj} \sum_d \pi_u(d)\, c_{d,j},
\]
where \(c_{d,j}\) is the label given by user \(d\) to pair \(j\).
The sign of \(s(q \mid u)\) determines which design is predicted to be preferred.

\subsection{Offline Evaluation on Preference Prediction}

We evaluate how well each method predicts a new user's preferences from only a few onboarding choices.

\ipstart{Methodology.}
We use leave-one-user-out cross-validation. In each of 20 folds, one participant serves as the new user and the other 19 form the training bank. We simulate onboarding by selecting \(N\) pairs from the training split with our EIG query policy and revealing the held-out user's labels for those pairs. Each method uses these observations to personalize its predictions on the same held-out test set. We report binary prediction accuracy averaged equally across users for \(N=0,1,2,4,6,8\).
We cap onboarding at eight comparisons to keep preference elicitation brief. This setup is similar in scale to YouTube Music's five-artist onboarding~\cite{hsu2024minimizing}; based on completion times in our prior preference collection study, eight comparisons correspond to roughly 2--3 minutes of active input, comparable to prior pairwise preference-elicitation interfaces that took 3.6 minutes on average~\cite{kuhlman2019evaluating}.

\ipstart{Baselines.}
We compare our method with other personalization alternatives that use the same set of onboarding samples. \emph{UIClip (pretrained)} is the base model we built upon, which provides a UI-specific shared evaluator without user-specific adaptation~\cite{wu2024uiclip}. \emph{UIClip + full tuning} updates the vision encoder, projection, and reward head from a source-trained initialization. \emph{UIClip + LoRA} instead applies rank-eight LoRA updates to the vision attention query and value projections while also updating the reward head~\cite{hu2022lora}. These two baselines test whether sparse feedback can directly adapt model parameters.
\emph{UIClip + VBPR} adapts visual Bayesian personalized ranking~\cite{he2016vbpr}. Prior users establish shared visual factors, and onboarding fits a factor vector for the new user. We use UIClip visual features rather than item identities so that the model can score previously unseen interfaces. \emph{KNN (nearest neighbors)} predicts from visually similar pairs among the new user's labeled onboarding examples.
\emph{GPT-5.2 Chat} infers the user's preferences directly from labeled UI examples in context. \emph{GPT + inferred rubric} first summarizes the onboarding example preferences into a personal decision guide using the dimensions identified in Section~\ref{sec:modeling_motivation}, then judges new pairs using the guide and the corresponding examples. Appendix~\ref{sec:prompts} provides the prompts.

\begin{figure}[!htbp]
    \centering
    \includegraphics[width=0.45\textwidth]{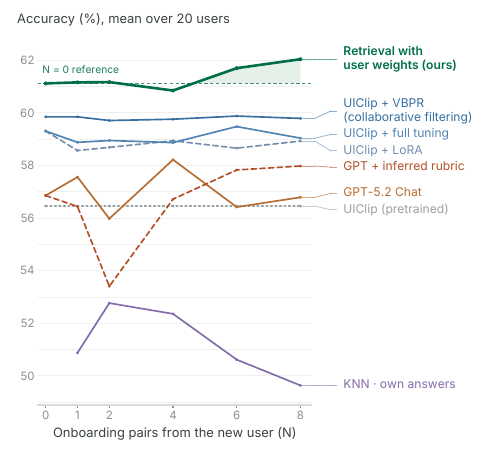}
    \caption{Preference prediction accuracy averaged over 20 held-out users. Our method has the highest mean accuracy at each evaluated onboarding budget. KNN requires at least one onboarding example, while pretrained UIClip does not use onboarding labels.}
    \label{fig:kcurve}
    \vspace{-10pt}
\end{figure}

\ipstart{Results.}
\label{sec:results}
Our method has the highest mean accuracy at each evaluated budget. At $N=8$, it reaches 62.05\%, compared with 59.80\% for VBPR, the strongest baseline. The paired difference is 2.24 percentage points (95\% CI [0.54, 3.79], participant bootstrap with the evaluated prompts fixed). Full tuning and LoRA reach 59.03\% and 58.92\%, respectively. GPT + inferred rubric reaches 57.98\%, compared with 56.78\% for direct GPT, while KNN remains near chance. Performance on prompting GPT-5.2 fluctuates across budgets without a consistent benefit from more examples. Prior work has found that few-shot prompting with UI critiques can improve general UI feedback~\cite{duan2024uicrit}, but our examples must convey an individual user's preferences rather than shared design quality. The inferred rubric recovers some performance at eight examples, but still falls short of methods that reuse prior users' labeled comparisons.

Our method starts at 61.12\% accuracy at $N=0$, compared with 56.46\% for pretrained UIClip. At $N=0$, user weights are uniform, so the model relies on preference information shared across prior users. Our dataset makes this signal directly usable because the same UI comparisons are labeled by multiple users, allowing the model to reuse population-level preference structure for a new user.
Performance increases gradually with larger onboarding budgets. Accuracy increases by 0.93 percentage points from \(N=0\) to \(N=8\). Across the six evaluated budgets, a descriptive linear fit has a positive slope of 0.11 percentage points per comparison and $R^2=0.61$. The gain is modest, but the trend suggests that a few pairwise judgments can refine an already useful shared prior. This is particularly useful in the sparse-feedback setting, where direct parameter adaptation through full tuning or LoRA remains less effective.
We intentionally evaluate a practical low-feedback setting because explicit onboarding imposes effort on new users. Recent preference learning studies have evaluated larger number of user input, including 10--100 onboarding samples~\cite{oh2024ample,padurean2025useralign}, although their tasks and feedback settings differ from ours. The offline results show that shared preference data provides a good basis for the initial prediction signal, while adaptive onboarding contributes additional evidence about the individual user. We next test whether this combination improves the interfaces that new users receive in an end-to-end generation setting.

\section{Online Study on GenUI Personalization}

Our offline evaluation shows that the adaptive retrieval model predicts held-out preferences more accurately than model adaptation and prompting baselines. That evaluation, however, uses a fixed set of interfaces and does not show whether the learned preference scores improve the interfaces users ultimately receive. We therefore conduct an online study with 12 new users, where interfaces are generated on the fly. We compare our full personalization pipeline with three baselines: generation without user information, generation conditioned on a user's written preference profile, and profile-conditioned generation reranked by a profile-conditioned LMM judge.

% \section{Online Study on GenUI Personalization}
% Our offline evaluations show that the adaptive, retrieval-based preference model works better than personalization through domain-specific model adaptation/tuning or prompting larger multimodal models. However, the offline evaluation only measures how well the model predicts preferences on a fixed test split from our static dataset, not whether the model can improve actual generation pipeline for new users. We thus run a follow-up online study with 12 new users to test whether our approach also improves personalized generation when interfaces are produced on the fly. We compare our full personalization pipeline against three baselines, generation with no user profile, generation conditioned on a user's text profile, and generation with the profile-conditioned judge as a reranker.

\subsection{Method}

\ipstart{Participants.}
We recruited 12 new users from our institution's mailing list, aged 25 to 32 (5 men, 7 women). All participants had used generative AI tools in their daily activities. Their UI design experience ranged from relevant university coursework to professional practice in industry. The asynchronous online study took approximately 45--60 minutes. Participants were compensated at the same hourly rate (\$15.55/hr) as in the prior data collection study, and our institution's IRB approved the study.

\ipstart{Procedure.}
Our study had two phases. In the onboarding phase, participants described their UI preferences as free text and rated 8 pairwise UI comparisons drawn from our preference dataset. In the arena phase, participants compared UIs generated on the fly from a set of screen prompts and personalized based on the onboarding responses, so each person compared a unique set of generated UIs.
The study began with a starting page with study instructions and a field for participants to enter their UI preferences as free text. To avoid biasing responses toward specific dimensions, we kept the guidance minimal and open-ended, asking participants to ``\textit{describe the kinds of interfaces you usually like as a user. You can mention things like simplicity, density, color style, or any other aspects you care about in UI screens.}'' Participants then rated 8 onboarding pairwise UI comparisons. The onboarding pairs were selected for each individual based on the query policy (Section~\ref{sec:modeling}), so each participant received a personalized set of onboarding questions. After a 10 to 15 minute wait for all conditions to generate, participants moved to the anonymized arena phase and compared UIs from four conditions. We choose the four conditions to separate the contribution of different forms of user information and selection. Prior work has personalized model outputs by conditioning on user profiles or preference descriptions~\cite{salemi2024lamp,ramos2024transparent}. We therefore compare generation without user information, generation conditioned on the user's written profile, reranking with the same written profile, and reranking with preferences learned from pairwise onboarding.
\begin{enumerate}[leftmargin=*, labelindent=0pt]
    \item \textbf{Zero-shot generation (Baseline).} An LLM generated a UI sample for a given prompt with no user information.
    \item \textbf{Profile-conditioned generation (Baseline).} An LLM generated a UI sample conditioned on the user's text profile. Profile-based personalization is also common in deployed systems; for example, Spotify maintains a taste profile to support recommendations, while Indeed Career Scout uses profile to personalize job recommendations.
    \item \textbf{Profile-conditioned generation + profile-conditioned LMM judge (Baseline).} An LLM generated samples conditioned on the user's text profile, and a profile-conditioned LMM judge selected the top choice.
    \item \textbf{Profile-conditioned generation + adaptive preference model (Ours).} An LLM generated samples conditioned on the user's text profile, and our adaptive preference model selected the top choice based on the onboarding preference labels.
\end{enumerate}
We chose GPT-5.2-chat as the base model for both generation and judge conditions based on its speed, cost, API quota limits, and output quality at the time we conducted the study. We produced 20 new UI prompts~\cite{wang2021screen2words} with the same method as our preference collection study, each from a unique screen type~\cite{leiva2020enrico} and with no overlap with prompts from the prior study. For the profile-conditioned conditions, we generated a shared pool of 16 samples per prompt and each condition selected its top choice from that same pool, which controls for randomness and reduces participant wait time. When any two conditions selected the same top sample, we assigned an automatic tie and removed the redundant comparison. All conditions used the same quality check prompt and annotation interface as the preference dataset collection study. 
(Appendix~\ref{sec:prompts}) Conditions were presented in randomized arena-style pairwise comparisons, with a maximum of 120 comparisons per participant (6 pairwise comparisons among the 4 conditions for each of the 20 prompts).

% \ipstart{Analysis.}
% We computed Bradley-Terry rating scores following standard arena methodology for comparing generative model outputs~\cite{zheng2023judging}. Higher scores indicate conditions that participants preferred more often. We report the median score with 95\% confidence intervals from bootstrap sampling as well as the average win rate for each condition.

\ipstart{Analysis.}
We analyze the online study from two perspectives. We first give a qualitative summary of users' free-text profiles and the optional justifications they provided during onboarding, which lets us characterize the preference dimensions participants specified. We then analyze the arena outcomes with Bradley-Terry scores following standard arena methodology~\cite{zheng2023judging}. Higher scores indicate conditions that participants preferred more often. We report bootstrap medians with 95\% confidence intervals, aggregate win rates, and direct head-to-head results. 
% Because the three profile-conditioned conditions all drew from the same candidate pool for each prompt, differences among them isolate selection quality rather than differences in raw generation.

\begin{figure}[!htp]
    \centering
    \includegraphics[width=\linewidth]{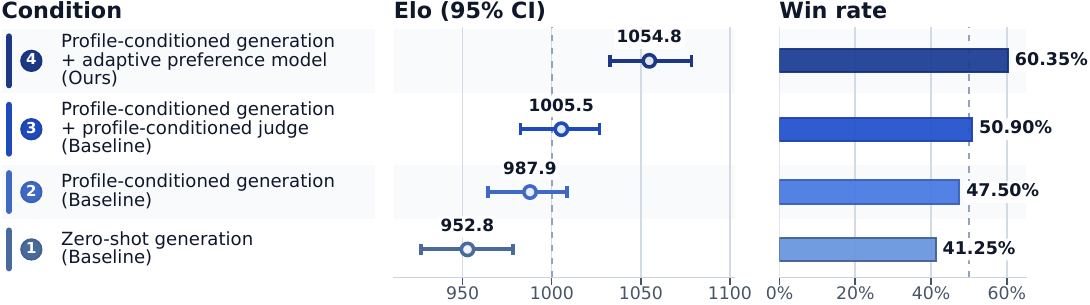}
    \caption{Arena-style ratings and win rates across four UI generation pipelines. Our adaptive method achieves the highest win rate and ranks first in 96.8\% of participant bootstrap resamples.}
    \label{fig:arena_results}
    \vspace{-12pt}
\end{figure}

% , and visual-versus-textual density

\subsection{Results}
\ipstart{Preference Profiles.}
The user free-text profiles were diverse in both length (median 28 words, range 8--72) and emphasis. Users described preferences along both aesthetic and information presentation dimensions, including cleanliness and simplicity, information hierarchy and action discoverability, modern or platform-specific styling, color tone, spacing and typography. Many profiles combined several considerations at once. Some users wanted expressive or visually distinctive screens as long as the UI remained easy to use, while others preferred iOS-like spacious layouts, darker themes, or more visual interfaces with meaningful icons.
In our study, all 12 participants received a unique sequence of onboarding queries, as the adaptive policy selected different comparisons based on each person's responses. The pairwise labels make the differences clearer than text profiles alone.
Qualitatively, users who described themselves as preferring expressive or modern screens still rejected candidates with weak contrast, while users who preferred minimal UIs sometimes chose denser layouts when the extra information made browsing easier. On the wallpaper gallery pair, preferences were evenly split. Some users valued denser browsing and visible filters, while others preferred cleaner framing, stronger contrast, or imagery that better fit the mobile aspect ratio. By contrast, the restaurant finder empty state drew unanimous preference for the version with clearer navigation and stronger contrast (See Supplementary). The examples illustrate trade-offs that broad profiles may leave unspecified.

% User profiles capture high-level taste, but pairwise labels expose the task-specific trade-offs that text alone misses.

% figure for each example? => appendix

\ipstart{Preferences on Generative UIs.}
As shown in Figure~\ref{fig:arena_results}, the pipeline powered by our preference model achieved the highest arena performance across the four conditions. Our method reached a 60.35\% aggregate win rate, compared with 50.9\% for the profile-conditioned judge, 47.5\% for profile-conditioned generation, and 41.25\% for zero-shot generation. 
Across bootstrap resamples of participants, our method ranked first in 96.8\% of runs.
The comparison against the profile-conditioned judge is especially important because both conditions used the same profile text and the same candidate pool. The difference lay in the final selection step. The profile-conditioned judge relied on the user's text profile, while our method used onboarding preference labels to rerank the candidates. The two selectors chose the same final UI in only 23 of 240 prompt-participant cases. Under this matched setup, our method performed better overall. 
The advantage also appeared across prompts. Aggregated over participants, our method outperformed zero-shot generation on 14 of 20 prompts, profile-conditioned generation on 16, and the profile-conditioned judge on 12. Free-text profiles helped when used to condition a judge, but the larger improvement came from the adaptive pairwise labels. With a few onboarding examples, the preference model reranked the candidate set in a way that matched each user better.

\vspace{-5pt}

\section{Discussion}

\subsection{Applying Learned Preferences}

Our online study applies the learned preference model to interfaces generated for a new request. Given generated candidates \(\mathcal{G}=\{g_1,\ldots,g_K\}\), the system selects \(g_u^*=\arg\max_{g_i\in\mathcal{G}} f_u(g_i)\), where \(f_u\) is a pointwise preference score derived from the pairwise model \(s(q\mid u)\) in Section~\ref{sec:modeling}. This same formulation can apply beyond generated candidates. For example, many applications expose configuration options such as color theme, text size, spacing, or information density. Given configurations \(\mathcal{V}=\{v_1,\ldots,v_M\}\), a system could render each configuration and select \(v_u^*=\arg\max_{v_i\in\mathcal{V}} f_u(\mathrm{render}(p,v_i))\). The preference model could also rank alternative presentations of the same content or application state. A system might choose among a compact list, a card-based layout, or a more detailed view based on the user's learned preferences. These examples illustrate how the same preference model could potentially support open-ended generation, finite configuration spaces, and interface adaptation during use.

\subsection{Preference Heterogeneity in Generative UIs}
Our results suggest complementary roles for shared preference data and individual feedback, consistent with personalized alignment research~\cite{li2025prefpalette,poddar2024vpl,li2026prefdisco}. Shared judgments support an initial predictor, while onboarding adapts to each indvidual's preferences.
In the online study, users described their preferences in free text before any comparison, and the profile-conditioned judge still performed worse than our learned selector. 
UI preferences may be partly tacit, or pairwise comparisons may pack latent information per label by forcing concrete trade-offs. Profiles and few-shot choices thus could provide complementary preference information.
These individual differences also raise a question about population-level modeling. Our dataset shows that some UI pairs split users nearly evenly, while others draw almost unanimous preference. A natural next step would be to predict the full preference distribution across a population rather than a binary choice for one user. Recent work on distributional reinforcement learning (RL) for language models has begun to explore how models can output calibrated distributions over multiple answers~\cite{puri2026reaching}. That direction could extend to visual preference modeling, though preference dimensions in UI design are more entangled and harder to verify than in the text and coding domains where distributional RL has been tested so far.

\subsection{Preference Uncertainty Within Individuals}
Preferences differ across people, but how stable are they within a single person? Standard preference modeling, from Dawid-Skene annotator models~\cite{dawid1979ml} to modern RLHF reward training~\cite{ouyang2022training}, typically assumes each annotator provides a deterministic label for a given comparison. The context of generative UI and design complicates that assumption. On contested pairs, some users switched choices across repeated or similar comparisons, and the interaction traces from our offline study revealed substantial variation in decision time, with some comparisons resolved in seconds and others requiring much longer deliberation. This is consistent with recent work on hidden context in preference learning~\cite{siththaranjan2023distributional}, which has shown that even a single annotator's judgments can shift depending on which evaluation criteria become salient at a given moment. In generative UIs and design, such context dependence is especially plausible. A user comparing two UIs may focus on layout in one moment and on color or action clarity in the next, and these criteria may favor different candidates.
Our current pipeline treats all onboarding labels equally and does not distinguish between confident and uncertain judgments. Yet the behavioral signals in our data suggest that not all labels carry the same weight. Quick, decisive comparisons likely reflect strong preferences, while slow, hesitant ones may reflect genuine ambivalence about a specific trade-off. Decision time, preference strength ratings, and consistency across similar pairs could all serve as indicators of per-label reliability. Accounting for this kind of within-person uncertainty is an open problem. In crowdsourcing, repeated labeling and annotator reliability models~\cite{raykar2010crowds,chen2013pairwise} have long been used to handle noisy labels, but these methods assume the noise is random rather than context-dependent. For personalizing generative UIs, where each user may provide only a handful of labels and the candidate space changes with every prompt, new approaches are needed that can estimate not just what a user prefers but how certain that preference is on any given comparison.

\subsection{Limitations and Future Work}
Our results suggest that generative UIs can be better personalized from a small number of samples, but our current work has several limitations that point to directions for future research. First, our preference dataset is limited to feedback from 20 users. In contrast to existing crowd-sourced generative UI and design benchmarks~\cite{webdev_arena_2025,design_arena_2025}, we chose to focus on people with a certain level of UI design knowledge. Yet a larger and more diverse sample could enable new analyses, such as whether users naturally form preference clusters around shared schools of taste given similar pairwise comparisons. Our dataset also intentionally asks all users to judge the same comparison pairs, which makes cross-user preference structure directly observable and reusable. In practice, users will encounter different prompts and generated candidates, so their feedback may have little or no pairwise overlap. Future work should study personalization from sparse, partially overlapping histories without requiring users to evaluate the same interfaces. A related open question is how much feedback a real user would provide in practice. Our study asked for 8 onboarding comparisons, but we do not yet know how many comparisons users would be willing to complete when working with a generative UI system in the wild, or how much preference signal could be collected passively from \textit{natural interactions} rather than \textit{explicit onboarding}. A larger prior dataset and more individual labels per user would both likely improve personalization performance.

Our modeling experiments explored only a subset of possibilities enabled by the dataset. We focused on the efficient few-shot setting, where the preference data serves as adaptation or prompting examples for a fixed model. Other directions remain open, including fine-tuning a decoder LMM on the full preference data or incorporating users' rationales as additional supervision. We expect our dataset to support these and other future explorations.

We also limit our experiments to a relatively stable set of generator families to keep the generated UI distribution consistent across our study. In practice, GenUI models change quickly, and a new model or model update may produce different interfaces for the same prompt. This can shift the distribution on which the preference model was learned and change the trade-offs users encounter, even if their underlying preferences remain stable. Future work should study how well learned preferences transfer across generator families and model versions, and whether the preference model needs to adapt as the generation distribution changes.

Our pipeline also personalizes by selecting from a shared candidate pool rather than by steering generation directly. This is a practical constraint. The generators that produce the best UI code today are large closed-source models that cannot be easily finetuned, and the quality gap between these models and open-weight alternatives remains large on benchmarks like WebDev Arena and Design Arena~\cite{webdev_arena_2025,design_arena_2025}. 
Our retrieve-and-rerank approach works around this constraint, but it means personalization can only choose among what the generator already produces. Earlier optimization-based systems like Supple~\cite{gajos2010supple} adapted UI layouts to individual ability and device constraints by making personalization part of the generation objective itself. Extending that idea to generative models, where the generation objective is conditioned on a learned preference function rather than filtered through post-hoc reranking, is a natural next step as open-weight models close the quality gap or closed-source APIs begin to support lightweight per-user adaptation.
The improvement from personalization in our offline preference study is modest under our practical few-shot setup. Offline pairwise accuracy is not necessarily a direct measure of downstream selection quality; prior work on reward models has shown that models with similar pairwise accuracy can produce different downstream outcomes~\cite{wen2025rethinking,zhou2025rmb}. We therefore evaluate the end-to-end effect separately. In our online study, users preferred interfaces personalized by our adaptive method more often than those from the baselines. Future work should study more directly how gains in offline prediction translate into downstream ranking quality and visible differences for users.

Finally, our experiments use single-page GenUIs that are interactive but lack backend services, persistent state, or multi-screen navigation. Future work could explore how interaction traces from more complete systems, where users perform real tasks, provide richer preference signals (e.g., hesitation during use~\cite{yang2026guide,peng2025designpref}). Such traces could also enable personalization~\cite{gajos2005arnauld,peng2019personaltouch,lai2019fitbird} and run-time accessibility improvements~\cite{mowar2024tab,mowar2025codea11y,peng2021say} that extends beyond single-page layout to multi-screen workflows and long-horizon interactions.

\vspace{-5pt} 
\section{Conclusion}
We have investigated whether individual preferences can be learned efficiently and used to generate better personalized interfaces in practice. To ground this question, we collected 12,000 pairwise preference labels from 20 users over the same 600 comparison pairs. The data show that people differ substantially in their preferences for generative UIs, and that the disagreement reflects differences in how individuals weigh aspects like density and contrast. Our offline experiments show that the most effective strategy combines a prior built from global data with adaptive example selection, outperforming models that simply use larger parameters. In an online study with 12 new users, a lightweight onboarding of eight pairwise comparisons was enough for our adaptive method to outperform generic and profile-conditioned prompting. We demonstrate that generative UIs can be better personalized with limited per-user feedback, and that global preference is valuable not simply as the final training objective but as a shared prior that individual adaptation builds on. Our work is a step towards generative UI systems that recognize there is no single alignment target that serves everyone, and that the diversity of human taste is not noise to average away but the very signal that makes meaningful personalization possible.

%%
%% The acknowledgments section is defined using the "acks" environment
%% (and NOT an unnumbered section). This ensures the proper
%% identification of the section in the article metadata, and the
%% consistent spelling of the heading.

% \begin{acks}
% To Robert, for the bagels and explaining CMYK and color spaces.
% \end{acks}

%%
%% The next two lines define the bibliography style to be used, and
%% the bibliography file.
\bibliographystyle{ACM-Reference-Format}
\bibliography{reference}

%%
%% If your work has an appendix, this is the place to put it.
% \clearpage

% \appendix

% \section{Model Hyperparameters}
% \label{sec:parameters}

% Here are the hyperparameteres we used for training and prompting VLMs:

% \begin{table}[htbp]
% \centering
% \caption{Training hyperparameters for UIClip.}
% \label{tab:hyperparams-uiclip}
% \begin{tabular}{ll}
% \toprule
% \textbf{Model} & UIClip \\
% \midrule
% \textbf{Learning rate} & $5 \times 10^{-4}$ \\
% \textbf{Schedule} & Cosine ($T_{\max}=24$, $\eta_{\min}=10^{-4}$) \\
% \textbf{Weight decay} & $10^{-2}$ \\
% \textbf{Batch size} & 64 \\
% \textbf{Gradient clipping} & 1.0 \\
% \textbf{Patience} & 5 \\
% \textbf{Margin multiplier} & 1.1 for ``much better'' labels \\
% \bottomrule
% \end{tabular}
% \end{table}

% \noindent For LMM (i.e., GPT-5), we set temperature = 1.0 when it serves as a generator and temperature = 0.0 when it serves as a judge.

% \section{Preference Annotation Interfaces}
% The annotation interface used in both our preference data collection and our online GenUI evaluation study.

% \begin{figure}[!htb]
%     \centering
%     \includegraphics[width=0.415\textwidth]{figures/ui_pref.pdf}
%     \caption{Our annotation UI with a screen description, two side-by-side variants, and four preference options.}
%     \label{fig:annot_ui}
%     \vspace{-12pt}
% \end{figure}

% \section{Prompts}

\clearpage
\onecolumn

\appendix

\section{Model and Baseline Configuration}
\label{sec:parameters}

Table~\ref{tab:model_config} summarizes implementation settings for our preference model that are not specified in Section~\ref{sec:modeling}. Table~\ref{tab:baseline_config} gives the main settings used for the offline baselines. Hyperparameters for adaptive baselines were selected using source-user validation; target test labels were not used for model or hyperparameter selection.

\begin{table}[htbp]
\centering
\small
\caption{Implementation settings for our preference model.}
\label{tab:model_config}
\begin{tabularx}{\linewidth}{@{}>{\bfseries}p{0.27\linewidth}X@{}}
\toprule
Component & Setting \\
\midrule
UI encoder &
Frozen UIClip image encoder (\texttt{biglab/uiclip\_jitteredwebsites-2-224-paraphrased}). \\

Residual adapter &
\(512\!\rightarrow\!1024\!\rightarrow\!512\) linear layers with GELU; residual scale initialized to \(0.5\); output embeddings are L2-normalized. \\

Training loss &
Ordinal NLL with hard-negative, swap-consistency, and ordinal calibration terms. \\

Optimization &
AdamW, learning rate \(10^{-3}\), weight decay \(10^{-5}\), at most 40 full-bank updates, validation after each update, patience 8. \\

User posterior &
Uniform initial prior; four-way response likelihood with \(\lambda=0.5\); posterior temperature \(T=2\). \\

Retrieval inference &
Top 100 pairs by direct/swapped similarity; exponential similarity weighting with temperature \(0.5\). \\
\bottomrule
\end{tabularx}
\end{table}

\begin{table}[htbp]
\centering
\small
\caption{Key settings for offline preference-prediction baselines.}
\label{tab:baseline_config}
\begin{tabularx}{\linewidth}{@{}>{\bfseries}p{0.24\linewidth}X@{}}
\toprule
Baseline & Setting \\
\midrule
UIClip pretrained &
Frozen pretrained UIClip evaluator with no user-specific adaptation. \\

Full tuning &
Updates the vision encoder, projection, and reward head from the source-trained initialization using the revealed onboarding labels. \\

LoRA &
Rank-8 LoRA updates on vision attention query and value projections, together with the reward head; base vision weights remain frozen. \\

VBPR &
Visual Bayesian personalized ranking over frozen UIClip features; onboarding fits the new user's latent factor. \\

KNN &
Retrieves visually similar pairs only from the new user's revealed onboarding examples for \(N>0\). \\

GPT-5.2 Chat &
Pairwise preference judge conditioned directly on the user's onboarding examples. Screenshots are resized to at most 512 pixels on the long side and encoded as JPEG. \\

GPT + inferred rubric &
Uses the same onboarding examples, first infers a preference guide over the dimensions identified in Section~\ref{sec:modeling_motivation}, then applies that guide and the original examples when judging new pairs. \\
\bottomrule
\end{tabularx}
\end{table}

% \section{Model Hyperparameters and Configuration}
% \label{sec:parameters}

% Table~\ref{tab:hyperparams-uiclip} lists the training hyperparameters for UIClip. For GPT-5.2-chat, we set temperature to 1.0 when it serves as a generator and temperature to 0.0 when it serves as a judge. All other parameters use default values.

% \begin{table}[htbp]
% \centering
% \caption{Training hyperparameters for UIClip.}
% \label{tab:hyperparams-uiclip}
% \begin{tabular}{ll}
% \toprule
% \textbf{Model} & UIClip \\
% \midrule
% \textbf{Learning rate} & $5 \times 10^{-4}$ \\
% \textbf{Schedule} & Cosine ($T_{\max}=24$, $\eta_{\min}=10^{-4}$) \\
% \textbf{Weight decay} & $10^{-2}$ \\
% \textbf{Batch size} & 64 \\
% \textbf{Gradient clipping} & 1.0 \\
% \textbf{Patience} & 5 \\
% \textbf{Margin multiplier} & 1.1 for ``much better'' labels \\
% \bottomrule
% \end{tabular}
% \end{table}

\begin{figure}[!htb]
    \centering
    \includegraphics[width=0.4\textwidth]{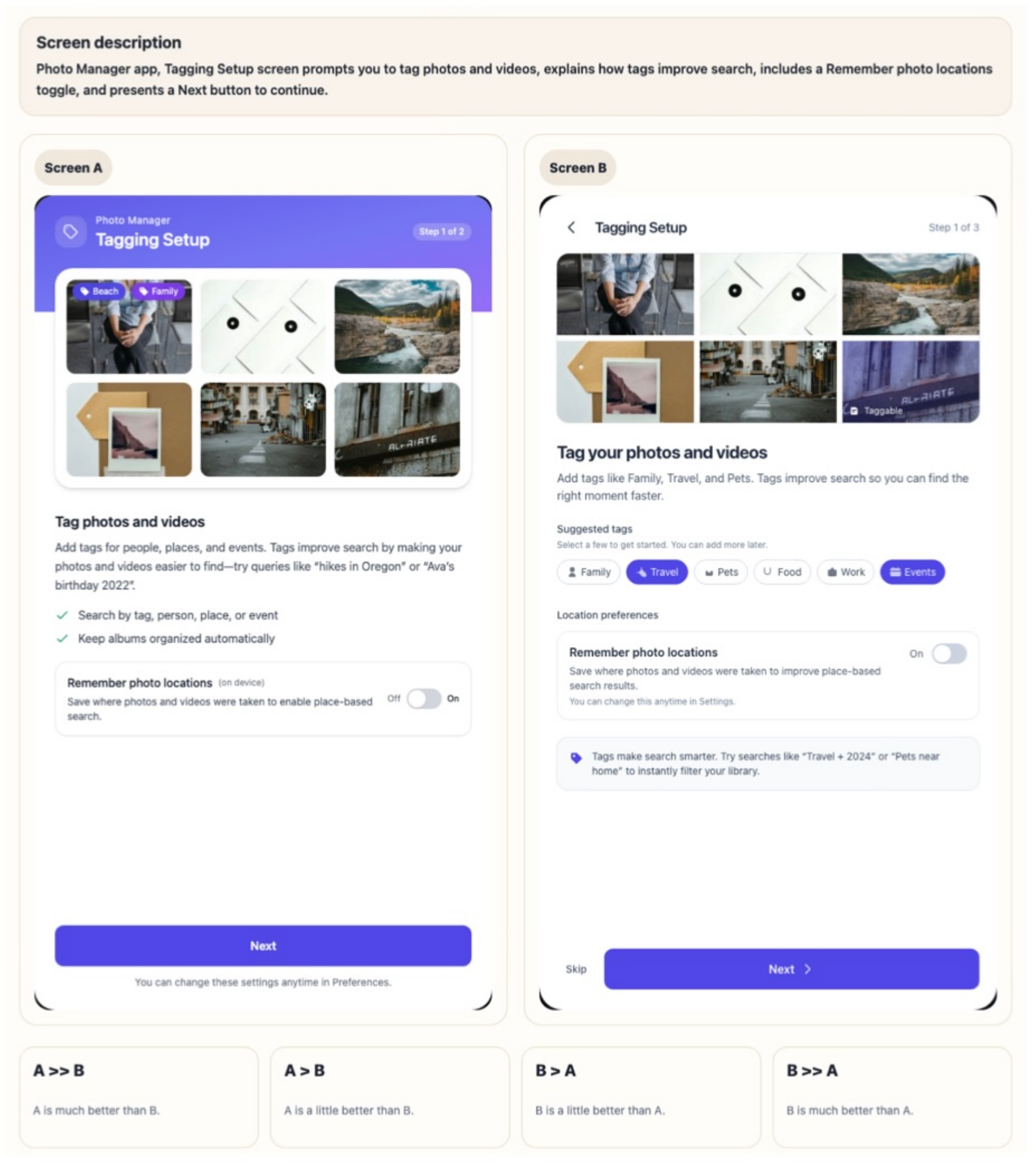}
    \caption{Our annotation UI with a screen description, two side-by-side variants, and four preference options.}
    \label{fig:annot_ui}
    \vspace{-8pt}
\end{figure}

\section{Annotation UI, Guidelines and Details}
\label{sec:annotation}

Figure~\ref{fig:annot_ui} shows the annotation interface used in both our preference data collection and our online evaluation study. Each trial presented a screen description, two side-by-side UI variants, and a four-point preference scale with no neutral option. The main instructions were as follows.

\ipstart{Preference ratings.}
For each pair, users selected one of four options using keyboard shortcuts (1: $A \gg B$, 2: $A > B$, 3: $A < B$, 4: $A \ll B$). They were instructed to focus on overall UI and design qualities such as layout, color, and style rather than placeholder content (e.g., generated text or unloaded images), and to treat flawed screens as starting points they would need to fix.

\ipstart{Rationale collection.}
In the follow-up study, users saw their earlier preference and wrote a rationale of at least 2--3 sentences. They were asked to give concrete visual and UI reasons (layout, color, hierarchy, typography, spacing, alignment, or missing components) and to avoid generic statements. Users could revise their earlier choice if their reasoning had changed.

\section{Prompts for LLMs and LMMs}
\label{sec:prompts}

We provide the prompt templates used for the GPT baselines and the online generation study. Braced fields denote values inserted at runtime.

\subsection{Offline Preference Judge}

The direct GPT baseline predicts a user's preference from their onboarding examples. At \(N=0\), the model receives only the target pair. For \(N>0\), the request first includes the user's onboarding pairs and choices, followed by the target pair.

\begin{tcolorbox}[
    breakable,
    enhanced,
    colback=blue!3,
    colframe=blue!55!black,
    boxrule=0.7pt,
    arc=2pt,
    left=6pt,
    right=6pt,
    top=5pt,
    bottom=5pt
]
\small
\textbf{System prompt}

\medskip
You are predicting pairwise UI preferences for one specific user. When onboarding examples are provided, they are previous choices from this same user. Use them to infer the user's preferences.

Consider clarity, readability, spacing and alignment, emphasis of primary actions, aesthetic balance, and fit to the screen prompt. Evaluate only visible features, avoid position bias, and do not allow ties.

\medskip
Return only:

\smallskip
\texttt{CHOICE\_4WAY: <A >> B | A > B | B > A | B >> A>}\\
\texttt{BINARY\_PREFERENCE: <A | B>}\\
\texttt{CONFIDENCE: <0.00-1.00>}\\
\texttt{REASONS:}\\
\texttt{- <short reason>}\\
\texttt{- <short reason>}
\end{tcolorbox}

\subsection{Inferred-Rubric Baseline}

The inferred-rubric baseline first summarizes the same onboarding examples into a user-specific preference guide. We constrain the guide to the four dimensions identified in our rationale analysis (Section~\ref{sec:modeling_motivation}): imagery use, readability versus visual polish, interface scope, and color theme. The resulting guide is fixed before test prediction and is provided to the same preference judge together with the original onboarding examples.

\begin{tcolorbox}[
    breakable,
    enhanced,
    colback=purple!3,
    colframe=purple!55!black,
    boxrule=0.7pt,
    arc=2pt,
    left=6pt,
    right=6pt,
    top=5pt,
    bottom=5pt
]
\small
\textbf{Preference-guide induction prompt}

\medskip
Infer this user's UI preferences from the onboarding examples.

Use only the following four dimensions:

\begin{enumerate}[leftmargin=1.6em, itemsep=1pt, topsep=3pt]
    \item \textbf{Imagery use:} less imagery versus more imagery.
    \item \textbf{Readability versus visual polish.}
    \item \textbf{Interface scope:} simplicity versus completeness.
    \item \textbf{Color theme:} light versus dark.
\end{enumerate}

For each dimension, infer the user's preferred direction when the examples provide evidence. Also indicate whether the dimension appears relatively more or less important to this user. If the evidence is mixed or insufficient, mark the dimension as unresolved.

Use only the visible onboarding examples and recorded choices. Do not invent user rationales, invisible functionality, additional preference dimensions, or numerical importance weights.
\end{tcolorbox}

\begin{tcolorbox}[
    breakable,
    enhanced,
    colback=purple!3,
    colframe=purple!55!black,
    boxrule=0.7pt,
    arc=2pt,
    left=6pt,
    right=6pt,
    top=5pt,
    bottom=5pt
]
\small
\textbf{Preference-guide context for test prediction}

\medskip
The following user preference guide was inferred only from the onboarding examples using the four preference dimensions above. Use the guide together with the original onboarding examples to predict the user's choice.

The onboarding examples remain the primary evidence. Treat the guide as a summary of those examples rather than an instruction that overrides them. When the guide provides no useful distinction for the target pair, rely on the examples and general visible UI quality.

\medskip
\texttt{USER\_PREFERENCE\_GUIDE:}\\
\texttt{\{preference\_guide\}}
\end{tcolorbox}

\subsection{Online Study Generation}

All online-study conditions use the same base UI generation instructions. Profile-conditioned conditions additionally receive the user's written preference profile. The generated screen must remain faithful to the requested task while using the profile to guide subjective visual choices.

\begin{tcolorbox}[
    breakable,
    enhanced,
    colback=orange!3,
    colframe=orange!65!black,
    boxrule=0.7pt,
    arc=2pt,
    left=6pt,
    right=6pt,
    top=5pt,
    bottom=5pt
]
\small
\textbf{Base UI generation prompt}

\medskip
You are an expert mobile UI developer.

\smallskip
\textbf{Goal.}
Output one complete standalone HTML document that renders one polished mobile UI screen for the provided request.

\smallskip
\textbf{Canvas.}
Wrap all screen content in a single \texttt{<div class="ui-canvas">...</div>} sized at \(628\times1118\) px. Keep the complete interface visible within the canvas with no clipping or scrolling.

\smallskip
\textbf{Design.}
Follow the requested task and content faithfully. Use clear hierarchy, readable typography, coherent spacing and alignment, appropriate contrast, and realistic mobile UI components. Include enough relevant content for the interface to feel complete, but do not add unrelated sections or shrink content simply to fit more elements.

\smallskip
\textbf{Tech.}
Use HTML, CSS, and JavaScript only. Tailwind may be loaded through CDN. Use inline SVG or a CDN icon library when needed.

\smallskip
\textbf{Images.}
Use either user-provided asset URLs or requested assets from pexel (based on the generated alt-texts).

\smallskip
\textbf{Accessibility.}
Use semantic elements when appropriate, associate labels with controls, provide meaningful image alt text, maintain readable contrast, and provide visible focus states.

\smallskip
\textbf{Return.}
Return valid HTML only, from \texttt{<!DOCTYPE html>} to \texttt{</html>}. Do not include Markdown, commentary, or code fences.
\end{tcolorbox}

For the profile-conditioned candidate pool, we append the following block:

\begin{tcolorbox}[
    breakable,
    enhanced,
    colback=orange!3,
    colframe=orange!65!black,
    boxrule=0.7pt,
    arc=2pt,
    left=6pt,
    right=6pt,
    top=5pt,
    bottom=5pt
]
\small
\textbf{User preference profile}

\medskip
Use the following profile as guidance for subjective interface choices such as visual tone, density, spacing, hierarchy, color palette, typography, and component styling.

Explicit task requirements, accessibility constraints, and output-format requirements take priority. Do not mention or quote the profile in the generated interface.

\medskip
\texttt{<user\_preference\_profile>}\\
\texttt{\{profile\}}\\
\texttt{</user\_preference\_profile>}
\end{tcolorbox}

The request then supplies the screen description and screen type:

\begin{tcolorbox}[
    breakable,
    enhanced,
    colback=orange!3,
    colframe=orange!65!black,
    boxrule=0.7pt,
    arc=2pt,
    left=6pt,
    right=6pt,
    top=5pt,
    bottom=5pt
]
\small
\textbf{Generation request}

\medskip
\texttt{Design brief:}\\
\texttt{\{screen\_description\}}

\medskip
\texttt{Screen type: \{screen\_type\}}

\medskip
Design one interface for this request. Do not mention candidate numbers, variants, internal testing, or comparisons in the UI. Preserve the main task, content, and primary user action.
\end{tcolorbox}

\subsection{Online Profile-Conditioned LMM Judge}

The profile-conditioned LMM judge selects one interface from the same profile-conditioned candidate pool used by our preference model. The judge receives the user's written profile but not the pairwise onboarding labels.

\begin{tcolorbox}[
    breakable,
    enhanced,
    colback=teal!3,
    colframe=teal!60!black,
    boxrule=0.7pt,
    arc=2pt,
    left=6pt,
    right=6pt,
    top=5pt,
    bottom=5pt
]
\small
\textbf{Profile-conditioned judge prompt}

\medskip
Choose the single best candidate screenshot for this user.

Use the following order:

\begin{enumerate}[leftmargin=1.6em, itemsep=1pt, topsep=3pt]
    \item Eliminate candidates that are visibly broken, difficult to read, poorly composed, or inconsistent with the requested screen.
    \item Among the remaining candidates, prefer stronger overall UI quality, including clear hierarchy, readable text, coherent spacing, alignment, contrast, and polished presentation.
    \item When multiple candidates are similarly strong, choose the candidate that best matches the user's written preference profile.
\end{enumerate}

Base the decision only on the requested screen, the user's profile, and the visible candidate screenshots.

\medskip
\textbf{User preference profile:}\\
\texttt{\{profile\}}

\medskip
Return only:

\smallskip
\texttt{\{"selected\_variant": "<number>",}\\
\texttt{ "reason": "<short reason>"\}}
\end{tcolorbox}

Our adaptive preference-model condition does not use an LMM judge. It ranks the same profile-conditioned candidate pool using the user's pairwise onboarding choices and the learned preference model.

\end{document}